# SEGMENTATION OF ALZHEIMER'S DISEASE IN PET SCAN DATASETS USING MATLAB

A. Meena[1], K. Raja[2]

[1]Research Scholar, Sathyabama University, Chennai, India
[2]Principal, Narasu's Sarathy Institute of Technology, Salem, India
kabimeena2@hotmail.com, raja_koth@yahoo.co.in

**Abstract**

Positron Emission Tomography (PET) scan images are one of the bio medical imaging techniques similar to that of MRI scan images but PET scan images are helpful in finding the development of tumors. The PET scan images requires expertise in the segmentation where clustering plays an important role in the automation process. The segmentation of such images is manual to automate the process clustering is used. Clustering is commonly known as unsupervised learning process of n dimensional data sets are clustered into k groups (k<n) so as to maximize the inter cluster similarity and to minimize the intra cluster similarity. This paper is proposed to implement the commonly used K- Means and Fuzzy C-Means (FCM) clustering algorithm. This work is implemented using MATrix LABoratory (MATLAB) and tested with sample PET scan image. The sample data is collected from Alzheimer's Disease Neuro imaging Initiative (ADNI). Medical Image Processing and Visualization Tool (MIPAV) are used to compare the resultant images.

## 1. Introduction

The most commonly used radiographic techniques are known as Computed Tomography (CT), Magnetic Resonance Imaging (MRI) and Positron Emission Tomography (PET). These technologies are major component techniques in diagnosis, clinical studies, treatment planning and are widely used for medical research. The motive of automatic medical image segmentation is to describe the image content based on its features. In recent years an ample of approaches has been proposed to segment the medical images according to its merits and limitations. The symmetry based approach is mathematically defined as a distance preserving transformation of the plane or space which leaves a given set of points unchanged and their respective features.

A PET is also known as PET imaging or a PET scan, is a type of nuclear medicine imaging. This scan image detects chemical and physiological changes related to metabolism. It uses a radioactive tracer element which is injected in the body and the tumors or cancers in the body are identified based on the movements of the tracer element [1]. This scan images are more sensitive than other image techniques such CT and MRI because the other imaging techniques only shows the physiology of the body parts where as the PET scan images shows the internal formation of tumors and cancer cells by means of the metabolism of the body parts [2].

This paper presents a study on the application of well known K-means and Fuzzy C-Means clustering algorithms. This algorithm is used to automate process of segmentation of the abnormal portion on the datasets classified by its type, size, and number of clusters [3]. The rest of the paper is organized as follows. Section 2 states the related work in this area. Section 3 explains the effect of Alzheimer's disease and section 4 describes the basic concept of clustering, section 5 covers the well known algorithm of K-Means and Fuzzy C-Means clustering. Section 6 presents the implementation method of these algorithms respectively. The result will be shown in section 7. Finally the concluding section is presented at last.

## 2. Related Work

Digital image processing allows an algorithm to avoid problems such as the build-up of noise and signal distortion occurs in analog image processing.

Michael J. Fulham et al., in the year 2002 stated that quantitative positron emission tomography provides the measurements of dynamic physiological and biochemical processes in humans. In 2003, Ciccarelli et al proposed a method sclerosis that disrupts the normal organization or integrity of cerebral white matter and the underlying changes in cartilage structure during osteoarthritis (Meder et al., 2006). Functional imaging methods are also being used to evaluate the appropriateness and efficacy of therapies such as Parkinson's disease, depression, schizophrenia, and Alzheimer's disease. Quantum dots (qdots) are fluorescent nano particles of semiconductor material is specially designed to detect the biochemical markers of cancer described by Carts-Powell, 2006.

Osama Abu Abbas in 2008 explained about the various clustering algorithm and its application based on the type of dataset used. In 2009, Stefan Kramer et al described the structured patient data for the analysis of the implementation of a clustering algorithm. The author expressed the medical research in dementia is to correlate images of the brain with other variables, for instance, demographic information or outcomes of clinical tests. In this paper, clustering is applied to whole PET scans.

Habib Zaidi et al [5] was applied a threshold value (T) to identify the the lesion foreground from a noisy background. In that threshold selection the PET image voxel are transformed into standardized uptake values. This paper expressed the unsupervised clustering method based on the fuzzy C means (FCM) algorithm. Here, the voxel might belong to more than one class based on its fuzzy membership functions [6, 7]. PET data constructed by four subsets and eight iterations. In [8], strongest kinetic activity is measured by active voxel with highest intensity. Then the active voxel is pre clustered. Ashburner et al. (1996) use the assumption that the PET data satisfy a Gaussian distribution, and Acton et al. (1999) use fuzzy clustering in [9, 10]. Xinrui Huang et al [11] described the combination of K-means and hierarchical cluster analysis was used to classify dynamic PET data.

## 3. Alzheimer's Disease

Alzheimer's Disease (AD) is a condition where the brain slowly goes down, as well as a serious loss of thinking ability in a person and cognitive impairment. PET scan image is mainly used for the Alzheimer's neurological disorder treatment and other kind of dementia. In 2006, there were 26.6million sufferers worldwide. Alzheimer's is predicted to affect 1 in 85 people globally by 2050[12,20]. Every five years after the age of 65, the risk of acquiring the disease approximately doubles, increasing from 3 to as much as 69 per thousand person years. Normally women are having a higher risk of developing AD [13]. Brain imaging using PET can validate Alzheimer's will typically show a reduction in glucose use in the cerebral cortex. The brain outermost layer is more responsible for many complex brain functions including memory, verbal and consciousness.

## 4. Clustering

Clustering is used to classify items into similar groups in the process of data mining. It also exploits segmentation which is used for quick bird view for any kind of problem. Unlike classification clustering and unsupervised learning do not rely on predefined classes and class labeled training examples. For this reason, clustering is a form of learning by observation, rather than learning by examples.

Many clustering techniques have been proposed over the years from different research disciplines. These techniques are used to perform with a given data and are being applied in an ample variety of interdisciplinary applications. In image segmentation, clustering is used to provide the updated centroid value according to the distance between the objects. The centroid calculation is described in fig.1.

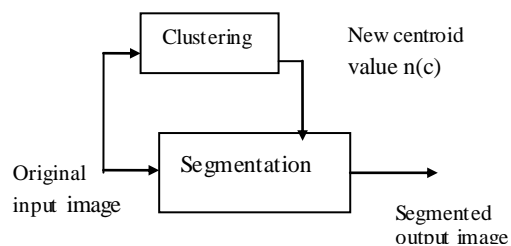

Fig. 1 Centroid calculation for image segmentation

## 5. Clustering algorithms

The most frequently known clustering algorithms are chosen to analyze. First the basic common cluster algorithm is known as K-Means and the second one is frequently known Fuzzy C-Means.

### 5.1 K-Means clustering algorithm

K-Means is a well known partitioning method. Objects are classified as belonging to one of k groups, k chosen a priori [3]. Cluster membership is determined by calculating the centroid for each group and assigning each object to the group with the closes centroid[14]. This approach minimizes the overall within-cluster dispersion by iterative reallocation of cluster members [15].

Pseudo code for centroid calculation

> **Step1**: Initialize / Calculate new centroid
> **Step2**: Calculate the distance between object and every centroid
> **Step3**: Object Clustering
> **Step4**: If any object moved from one cluster to the other, go to step1 or Stop

Pseudo code for image segmentation

> **Step1**: Initialize centroids corresponding to require number of clusters
> **Step2**: Calculate original centroid (Call K- Means)
> **Step3**: Calculate the mask
> **Step4**: Do the segmentation process

### 5.2 Fuzzy C-Means clustering algorithm

In 1969, Ruspini has given the idea of using fuzzy set theory for clustering. The first specific formulation of Fuzzy C-Means (FCM) is credited to Dunn [17]. But its generalization and current framing is designed by Bezdek [18].

In fuzzy set theory, every element in the universe belongs to a varying degree to all sets defined in the universe. But in fuzzy clustering, objects are not classified as belonging to one and only cluster, but instead, they own a degree of membership with each of the clusters. FCM provides hyper spherically-shaped well separated clusters accurately.

Pseudo code for Fuzzy C-Means

> **Step 1**: Initialize the following factors
> - Number of clusters
> - Assign centroid
> - Number of iterations
> - Termination parameters
> - Fuzziness factor
>
> **Step 2**: Calculate / update membership values($\mu_{ik}$)
> - Calculate distance($d_{ik}$)
> - Membership values are to be calculated using calculated distances
>
> **Step 3**: Update centroids
> **Step 4**: Find Objective function($J_r$)
> **Step 5**: If $J_r <> J_{r-1}$ then goto step 2 or stop

## 6. Implementation Method

The two basic algorithms K-Means and FCM is tested in MATLAB 7.0.1 with synthesis PET scan image datasets.

### 6.5 MATLAB Applications

MATrix LABoratoty(MATLAB) is widely used for implementing algorithms in numerical environment. The image processing tool box has a stable, well supported set of software tools for wide range of digital image processing and segmentation. The major applications are intensity transformation, image restoration, registration, image data compression, morphological image processing, regions and boundary representation and description. However, some limitations are listed in MATLAB such as its low processing speed and wasteful use of memory [19].

### 6.6 K- Means Clustering

The given image is loaded in to MATLAB. First the number of clusters is assigned. Then the centroid (c) initialization is calculated as follows

$$c = (1:k) * m / (k+1) \qquad (1)$$

where the double precision image pixel in single column (m) value and number of centroid (k) is used to calculate the initial centroid value.

The calculation of distance (d) between centroid and object is derived from

$$d = abs(o(i) - c) \qquad (2)$$

Equation (2) o(i) is known as one dimensional array distance. Using that value, new centroid is calculated in equation 3.

$$nc(i) = sum(a.*h(a)) / sum(h(a)) \quad (3)$$

where the value object clustering function (a) and non zero element obtained from object clustering h(a) is used to compute the new centroid (nc) value. The resultant new centroid value is used for masking creation and then the image segmentation.

### 6.7 Fuzzy C-Means clustering

The input image is loaded in to MATLAB. The number of clusters, the number of iterations and the membership value is assigned in prior. Table 1 is defined the cluster parameters used in the FCM clustering algorithm.

Table.1 The cluster parameters

| Parameter | Description |
|---|---|
| ncluster | Number of clusters |
| $x_k$ | $k^{th}$ data |
| center | Number of centroid |
| n | Number of data available |
| m' | Fuzziness factor |
| $d_{ik}$ | Distance between $k^{th}$ data and $i^{th}$ centroid value |
| $J_r$ | $J_r$ is the objective function value of $r_{th}$ iteration |
| mf | Membership function value |
| **$\mu_{ik}$** | Membership value of $k^{th}$ data and $i^{th}$ centroid |

$$d_{ik}=abs((ones(ncluster,1)*imgv') - repmat(center,[1\ imgsiz]))+0.000001 \quad (1)$$

Equation (1) is used to calculate the distance between the $k^{th}$ data and $i^{th}$ centroid value

$$\mu_{ik}= dik.\hat{}(-2/(expo-1))./(ones(ncluster,1) *sum(dik.\hat{}(-2/(expo-1)))) \quad (2)$$

$$mf= \mu_{ik}.\hat{}expo \quad (3)$$

Membership value $\mu_{ik}$ is computed in equation 2 and equation 3.

$$center=(mf*imgv)./((sum(mf'))') \quad (4)$$

The revise centroid is evaluated in equation 4.

$$obj(i)=sum(sum((dist.\hat{}2).*mf)) \quad (5)$$

$$abs(obj(i)-obj(i-1))<0.00001 \quad (6)$$

The equation 5 and equation 6 is used to compute the value of objective function and to give the exit factor related to the given iteration.

## 7. Result

The obtained image from MATLAB using K-Means is shown in fig.2 Fig.2a represents the original PET scan image and fig. 2b shows the segmented form of PET scan image having Alzheimer's disease.

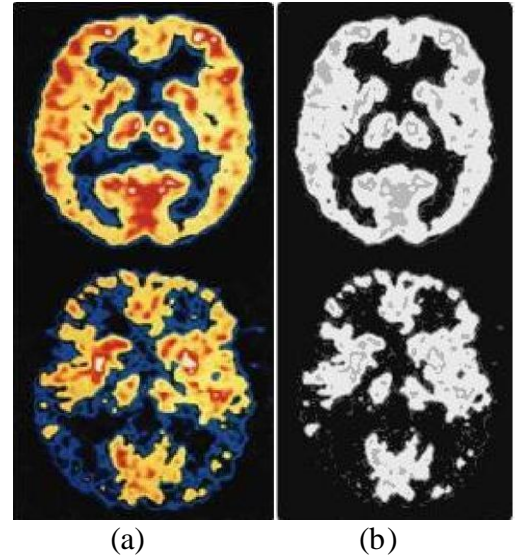

(a)        (b)

Fig. 2 K- Means image segmentation (a) Original Image, (b) Segmented Image

Fig. 3 is shown the resultant image obtained from FCM. Here, randomly the number of iteration is selected as three and the fuzziness factor as two. The below figure is shown all the three iteration using FCM. During the calculation, the histogram leads some spatial information loss. Instead of doing histogram vectorisation from vector to do the matrix format is provided the best solution.

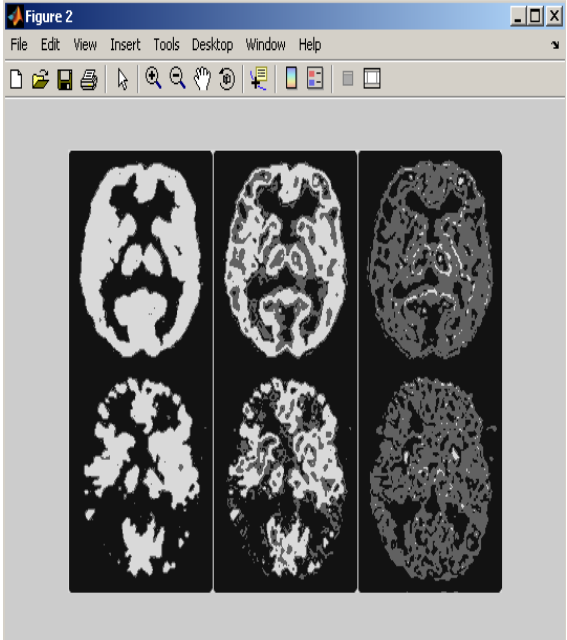

Fig. 3 Fuzzy C-Means image segmentation for the number of iteration as three

Table 2. Statistical analysis using MIPAV

| S. No | Parameter | Result obtained from | |
|---|---|---|---|
| | | K- Means | FCM |
| 1. | Average voxel intensity | 86.0916 | 76.4097 |
| 2. | Standard Deviation | 92.0758 | 42.8731 |
| 3. | Coefficient of variance | 106.951 | 56.109 |

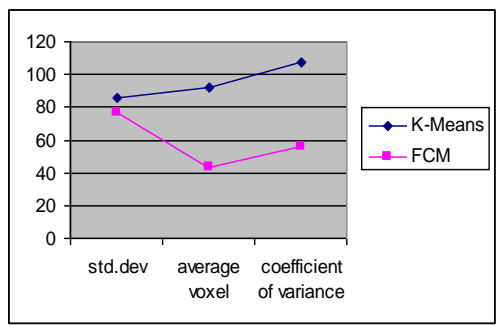

Fig.4 comparison between K-Means and FCM

In Fig. 4 shows an implementation effect of K-Means and FCM using MATLAB.

## 8. Conclusion and Future work

The bio medical imaging techniques have been prominently used for the clinical purpose such that the anatomy and the physiology of the internal parts can be monitored [8]. PET scan is a bio medical nuclear imaging techniques provide a solution for abnormal cells. The segmented image provides the clear picture about the affected portions. This paper explained the basic K-Means algorithm and FCM. Both algorithms are implemented in MATLAB. The cluster allocation is based on less distance and high membership value in FCM. In future the globally optimal segmented image with exact CPU utilization is to be found and this approach should be verified in real PET image datasets.